\documentclass[letterpaper, 10 pt, journal, twoside]{IEEEtran}

\usepackage{graphicx}
\usepackage{multirow}
\usepackage{siunitx}
\usepackage[caption=false,font=normalsize,labelfont=sf,textfont=sf]{subfig}

\IEEEoverridecommandlockouts

\begin{document}

\title{DIJIT: A Robotic Head for an Active Observer}

\author{
Mostafa~Kamali~Tabrizi,
Mingshi~Chi,
Bir~Bikram~Dey,
Kelly~Yuan,
Markus~D.~Solbach, \\
Yiqian~Liu,
Michael~Jenkin, \IEEEmembership{Senior Member, IEEE},
and~John~K.~Tsotsos, \IEEEmembership{Life Fellow, IEEE}
\thanks{Manuscript received: September 21, 2025; Revised January 29, 2026; Accepted March 24, 2026.}
\thanks{This paper was recommended for publication by Xinyu Liu upon evaluation of the Associate Editor and Reviewers’ comments.}
\thanks{
All authors were with the Department of Electrical Engineering and Computer Science, York University, 4700 Keele Street, Toronto, Ontario, Canada, at the time they were working on this research.
Email addresses:{
Mostafa Kamali Tabrizi: {\tt\footnotesize mkamali@yorku.ca},
Mingshi Chi: {\tt\footnotesize mingshic@yorku.ca},
Bir Bikram Dey: {\tt\footnotesize bir.b.dey@gmail.com},
Kelly Yuan: {\tt\footnotesize kellyuan@yorku.ca},
Markus D. Solbach: {\tt\footnotesize solbach@yorku.ca},
Yiqian Liu: {\tt\footnotesize yql@yorku.ca},
Michael Jenkin: {\tt\footnotesize michael.jenkin@lassonde.yorku.ca},
and John K. Tsotsos: {\tt\footnotesize tsotsos@yorku.ca}.
}
}
\thanks{Digital Object Identifier (DOI): 10.1109/LRA.2026.3682980}
}

\markboth{IEEE Robotics and Automation Letters. Preprint Version. Accepted April, 2026}
{Kamali Tabrizi \MakeLowercase{\textit{et al.}}: DIJIT: A Robotic Head for an Active Observer} 

\maketitle

\begin{abstract}

We present DIJIT, a novel binocular robotic head expressly designed for mobile agents that behave as active observers.
DIJIT's unique breadth of functionality enables active vision research and the study of human-like eye and head-neck motions, their interrelationships, and how each contributes to visual ability.
DIJIT is also being used to explore the differences between how human vision employs eye/head movements to solve visual tasks and current computer vision methods. 
DIJIT's design features nine mechanical degrees of freedom, while the cameras and lenses provide an additional four optical degrees of freedom. 
The ranges and speeds of the mechanical design are comparable to human performance.
DIJIT attains 85\% of the peak human saccade speed.
Our design includes the ranges of motion required for convergent stereo, namely, vergence, version, and cyclotorsion.
Here, we present DIJIT and some aspects of its performance.
We also present a novel method for saccadic camera movements, using a direct relationship between camera orientation and motor values.
The resulting saccadic camera movements are close to human movements in terms of their accuracy, with 1.17$^\circ$ and 1.14$^\circ$ mean error for the left and right cameras, respectively.

\end{abstract}


\section{Introduction}

\noindent
\IEEEPARstart{R}{obotic} and human visual systems differ significantly; robotic vision systems often utilize fewer degrees of freedom (DOF) and are typically designed for a small number of specific tasks, while human vision offers many more DOF in terms of camera/eye control and is intended for a wider range of tasks. 
Exploring the differences between computer and machine vision methods can provide insights into how eye and head movements aid in solving visual tasks.

In humans, vision tasks employ active vision. 
Eye fixation direction is actively controlled as part of visual task solutions.
Active vision in robotics has become a research paradigm inspired in part by biological vision systems~\cite{bajcsyRevisiting, findlayActiveVision2003}.
It has been shown to have benefits for various tasks, including Simultaneous Localization and Mapping (SLAM)~\cite{placedSurveyActiveSimultaneous2023} and object recognition~\cite{guan2024active}.
Typically, active vision machine systems are monocular or utilize a fixed binocular stereo camera pair on a pan-tilt system.
In humans, stereo vision is more than just two fixed eyes rotating on a pan-tilt neck; rather, each eye is highly adjustable to its viewing conditions, and its movements are controlled by a complex set of muscles that provide three rotational DOF to each eye, and the head itself is mounted on a controllable neck. 

\begin{figure}[!t]
    \centering
    \subfloat[]{\includegraphics[height=1.5in]{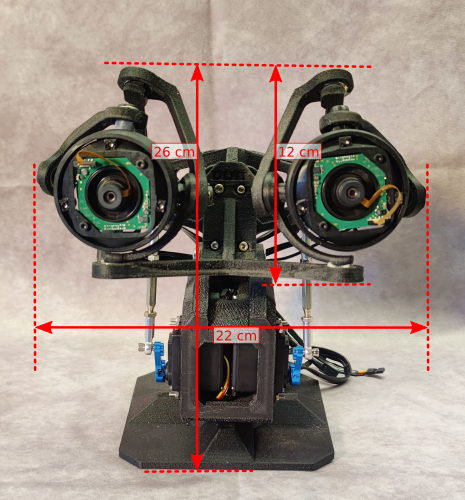}%
    \label{fron_view}}
    \hspace{0.1mm}
    \subfloat[]{\includegraphics[height=1.5in]{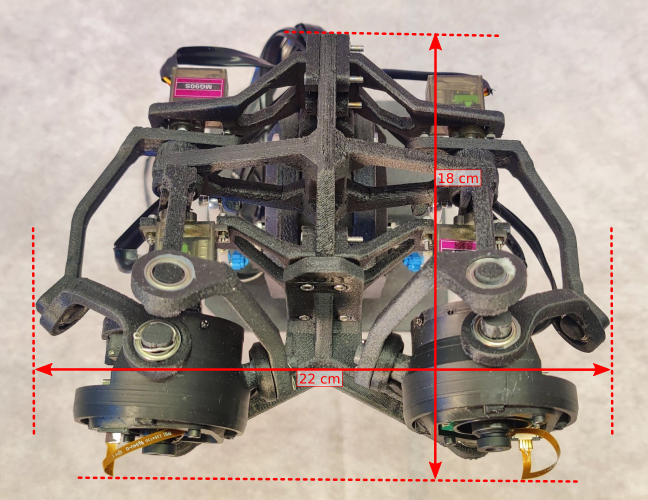}%
    \label{top_view}}
    \caption{
    The DIJIT robotic head from different viewpoints.
    (a) Front view, including dimensions for the width, height without the neck, and height with the neck.
    (b) Top view, including dimensions for the width and the depth without the neck.
    }
    \label{fig_head_photos}
\end{figure}

To explore the versatility that the human visual system has over active computer vision systems, a human-like binocular robotics platform is needed.
The DIJIT head (see Fig.~\ref{fig_head_photos}) mirrors a wide range of the properties of the physical human vision system, and was designed to address this need.

Biomimetic systems in robotics, including stereo heads that mimic human binocular stereoscopic systems, are a popular area of study for improving existing systems and evaluating visual theories.
The development of active stereo heads has a long history, tracing back to the early 1970's~\cite{tenenbaum_laboratory_1971, moravec_stanford_1983}. 
To have a robotic system that fully encompasses the human visual system, many properties must be considered, including the DOF of the cameras, the kinematic structure including the inter-camera baseline (the distance between the two camera centers), and the speeds and ranges of camera motion. 
The size and weight of the device come into play when performing a task that involves movement in 3D space, especially when mounted on a mobile robotic platform.
Existing robotic heads fall short in one or more of these aspects, often with designs relying on large and heavy metal frames and incorporating large inter-camera baselines.
Very few existing heads encompass all the necessary DOF, and even fewer authors detail their performance.
To the best of our knowledge, DIJIT is the first robotic head with baseline and mechanical DOF similar to those of humans (three DOF for each camera and three for the neck), in addition to four optical DOF for each camera.

Here, DIJIT's ability to perform similarly to humans is evaluated in terms of its ability to execute a saccade, the most frequent voluntary human eye movement, occurring 2-3 times per second~\cite{saccadeFrequency}, making it essential for biological active vision. 
DIJIT attains over 85\% of the peak saccade speed relative to human performance and has an accuracy comparable with human performance.
A supplementary video demonstrates DIJIT in operation.

The remainder of this paper is organized as follows.
Section~\ref{related_work} reviews the characteristics of existing robotic heads.
Section~\ref{design} describes DIJIT's mechanical, hardware, and software design.
Section~\ref{sec:saccade} describes an approach to saccade control using DIJIT. 
This approach takes advantage of the mechanical and visual properties of the head.
Section~\ref{experiments} provides experimental results on executing saccades on DIJIT.
Finally, Section~\ref{sec:review} summarizes the paper and describes ongoing and planned future work with DIJIT.

\section{Previous Work} \label{related_work}

\subsection{Stereo Head Designs}

\noindent For an earlier review of stereo head designs, see~\cite{ikeuchi_evolution_2021}.
Table~\ref{tab:heads} updates the set of stereo heads reviewed in~\cite{ikeuchi_evolution_2021} and summarizes some of their characteristics.
The heads reviewed here were chosen based on the design goal to be ``human-like''.
Most existing human-like robotic heads can be categorized into one of four groups based on their kinematic structure:

\begin{enumerate}

    \item {Systems that use fixed stereo camera geometry; such as ATLAS~\cite{atlas_percep} and THORMANG~\cite{thormang};} \label{cat_1}

    \item {Systems that utilize cameras that pan individually and utilize a common tilt; such as Manfredi~\cite{manfredi_implementation_2006}, the COG Project~\cite{goos_cog_1999}, iCub~\cite{icub_head}, and Tombatossals~\cite{antonelli_learning_2015};} \label{cat_2}
    
    \item {Systems that utilize cameras that pan and tilt individually; such as OREO~\cite{huber_oreo_2018}, KTH~\cite{KTH}, ESCHeR (ETL)~\cite{escherETL}, Agile~\cite{samson_agile_2006}, and Zhang~\cite{xiaolinzhang}; and,} \label{cat_3}
    
    \item {Systems that utilize cameras that pan, tilt, and roll independently; such as ETL-Humanoid~\cite{etl_humanoid_cognition}, Robot Bionic Eyes~\cite{robotbionichead}, TRISH~\cite{milios_design_1993}, MAC-EYE~\cite{cannata}, and Van Opstal~\cite{vanOpstal}.} \label{cat_4}
    
\end{enumerate}

\noindent
DIJIT falls in the fourth category; it is equipped with a full three rotational DOF for each of its two cameras.
This is augmented by a neck providing an additional three mechanical DOF.
Considering the heads in Category~\ref{cat_4}, only ETL-Humanoid~\cite{etl_humanoid_cognition} and Robot Bionic Eyes~\cite{robotbionichead} also have three DOF in the neck.
Robot Bionic Eyes~\cite{robotbionichead} does not provide optical DOF and has lower resolution cameras.
TRISH~\cite{milios_design_1993}, while having very precise motors, has only one DOF in the neck, and does not have a human-like inter-camera baseline.
MAC-EYE~\cite{cannata} focused on mimicking the muscles of the human eye with a sphere actuated by four independent tendons and motors.
However, the robot does not include a neck \cite{cannata_saccade}.
The Van Opstal robot~\cite{vanOpstal} uses six independent tendons that resemble human eye muscles; however, only a single camera has been constructed, with the experiments reported in simulation.

Of all the heads reviewed, only iCub~\cite{icub_head}, OREO~\cite{huber_oreo_2018}, and Agile~\cite{samson_agile_2006} have a baseline that is close to that of humans. However, iCub~\cite{icub_head} only implements three DOF in total in the two cameras, and many experiments on the real robot use fixed camera orientation. OREO~\cite{huber_oreo_2018} lacks an individual camera roll and does not report on saccade performance, focusing instead on speed.
In Agile~\cite{samson_agile_2006}, each camera is bulky, 20.5cm$\times$20cm (W$\times$H).

\begin{table}[!t]
    \renewcommand{\arraystretch}{1.3}
    \caption{Comparison Between Different Robotic Heads
    \label{tab:heads}}
    \centering
    \begin{tabular}{l|ccc}

    \hline
    & \multirow{2}{*}{DOF$^{(1)}$} & \multirow{2}{*}{Baseline} & Motion Range$^{(2)}$ \\
    \multicolumn{1}{c|}{Head} & \multirow{2}{*}{(Eyes, Neck)} & \multirow{2}{*}{(mm)} & (eye pan$^\circ$; eye tilt$^\circ$) \\
    &  &                           & (head pan$^\circ$; head tilt$^\circ$) \\
    \hline
    \multirow{3}{*}{Human}                          & \multirow{3}{*}{(6, 3)}                        & \multirow{3}{*}{45-80}                     & ($\pm$45; --47, +28)      \\
                                                    &                               &  & ($\pm$60; --30, +60)      \\
                                                    &~\cite{MarkusNeckDOF}          &~\cite{baselineLength}     &~\cite{Lee2019-rz, Head_ROM} \\
    \hline
    \multirow{2}{*}{DIJIT}                                           & \multirow{2}{*}{(6, 3)}                        & \multirow{2}{*}{115}                       & ($\pm$40; $\pm$46)      \\
    & & & ($\pm$135; --40, +38)      \\
    \hline
    ETL-Humanoid~\cite{etl_humanoid_cognition}      & (6, 3)                        & N/R$^{(3)}$               & N/R           \\
    \hline
    \multirow{2}{*}{RBE$^{(4)}$~\cite{robotbionichead}}              & \multirow{2}{*}{(6, 3)}                        & \multirow{2}{*}{118}                       & (100; 100)    \\
    & & & (90; 90) \\
    \hline
    \multirow{2}{*}{TRISH~\cite{milios_design_1993}}                 & \multirow{2}{*}{(6, 1)}                        & \multirow{2}{*}{320}                       & ($\pm$35; $\pm$45)      \\
    &&&($\pm$80; ---)\\
    \hline
    \multirow{2}{*}{MAC-EYE~\cite{cannata}}                          & \multirow{2}{*}{(6, 0)}                        & \multirow{2}{*}{N/R}                       & ($\pm$55; $\pm$55)    \\
    &&& (---; ---) \\
    \hline
    \multirow{2}{*}{Van Opstal~\cite{vanOpstal}}                     & \multirow{2}{*}{(6, 0)}                        & \multirow{2}{*}{One eye$^{(5)}$}           & ($\pm$50; $\pm$50)      \\
    &&&(---; ---)\\
    \hline
    \multirow{2}{*}{OREO~\cite{huber_oreo_2018}}                     & \multirow{2}{*}{(4, 3)}                        & \multirow{2}{*}{58}                        & ($\pm$41; --48, +41)      \\
    &&&($\pm$90; --55, +80)\\
    \hline
    KTH~\cite{KTH}                                  & (4, 2)                        & 150-400                   & N/R           \\
    \hline
    \multirow{2}{*}{ESCHeR (ETL)~\cite{escherETL}}                   & \multirow{2}{*}{(4, 1)}                        & \multirow{2}{*}{180}                       & (100; 90)    \\
    &&&(200; ---)\\
    \hline
    \multirow{2}{*}{Agile~\cite{samson_agile_2006}}                  & \multirow{2}{*}{(4, 0)}                        & \multirow{2}{*}{50-550}                    & ($\pm$40; $\pm$40)      \\
    &&& (---; ---) \\
    \hline
    Zhang~\cite{xiaolinzhang}                       & (4, 0)                        & N/R                       & N/R           \\
    \hline
    \multirow{2}{*}{Manfredi~\cite{manfredi_implementation_2006}}    & \multirow{2}{*}{(3, 4$^{(6)}$)}                & \multirow{2}{*}{N/R}                       & ($\pm$45; --25, +53)      \\
    &&&($\pm$108; --28, +32)\\
    \hline
    \multirow{2}{*}{COG~\cite{cog}}                                  & \multirow{2}{*}{(3, 3)}                        & \multirow{2}{*}{$\sim$125}                      & (120; 60)     \\
    &&&(180; N/R)\\
    \hline
    \multirow{2}{*}{iCub~\cite{icub_head}}                           & \multirow{2}{*}{(3, 3)}                        & \multirow{2}{*}{$\sim$68}                       & (90; 80)      \\
    &&&(110; 90)\\
    \hline
    Tombatossals~\cite{antonelli_learning_2015}     & (3, 1)                        & 270                       & N/R           \\
    \hline
    
    \end{tabular}
    
    \begin{flushleft}
    \footnotesize
    $^{(1)}$ Only rotational DOF are considered.
    $^{(2)}$ Motion extent is specified with respect to the axis where available; otherwise, total range is given.
    $^{(3)}$ N/R stands for Not Reported.
    $^{(4)}$ Robot Bionic Eyes
    $^{(5)}$ Van Opstal~\cite{vanOpstal} is only a prototype of one eye; it does not have a full head, and the experiments are conducted in simulation; however, we consider six DOF for the eyes, assuming a full head.
    $^{(6)}$ Manfredi~\cite{manfredi_implementation_2006} has four DOF in the neck because its neck has two tilts at two different levels, in addition to pan and roll.
    \end{flushleft}
     
\end{table}

\subsection{Saccade Methods}

\noindent
Just as humans use saccades to fixate different locations of interest in the visual scene, robotic heads must decide ``where to look next'' and execute that motion.
In biological systems, these motions are known as saccades or saccadic motions.
We briefly review existing methods proposed for saccadic motion for robotic heads. 
We concentrate on how to actually redirect an individual camera, rather than on the problem of where to look next.
For a review of where the head should attend, see~\cite{SaliencyModels}.

In biological systems, saccades often take place in multiple phases.
A primary saccade drives the eye to the direction of interest, and this can be followed by a corrective saccade to refine the motion executed in the primary saccade.
See~\cite{corrective_saccade} for a review on corrective saccades.
Machine systems can also benefit from a multi-stage saccade strategy to deal with errors in mapping from camera measurements to motor commands to backlash in the mechanical motion system used.
Here, we concentrate on primary saccades, but also touch on corrective saccades as appropriate.

Some saccade execution methods (e.g.,~\cite{antonelli_learning_2015, vanOpstal}) model the kinematics, dynamics, and/or optics of the robot, typically requiring substantial time for training and/or execution;
Other methods (e.g.,~\cite{antonelli_learning_2015, schenck_robot_2013, wilson_integration_2015}) train neural networks to learn saccade strategies, which requires extensive data collection and training.
The novelty of our approach is that it uses homography mappings between images collected in a fast calibration data collection stage to assign appropriate motor values to each pixel $\mathbf{t}$ in the viewed image, so that when the motors move to those values, the camera fixates on the point corresponding to pixel $\mathbf{t}$.
The approach presented for COG~\cite{maes_self-taught_1996} is the most similar to our approach in terms of assigning motor values to pixels in the image, where they utilize an online learning method to create a saccade map.    
However, they report 90 minutes of processing time for calibration, limiting the applicability of the approach; whereas, our method takes less than three minutes for calibration of each camera.
Our method is faster than other reported methods, and in terms of accuracy, it is more accurate or comparable to other methods, see Section~\ref{experiments}.

\section{DIJIT Design} \label{design}

\noindent
Here, we describe the mechanical design, hardware components, and ROS interface of DIJIT. 
The 3D part models, parts list, and software code are all available online under the MIT license (https://gitlab.nvision.eecs.yorku.ca/robots/dijit-pub).

\subsection{Mechanical Design}

\begin{figure}[!t]
    \centering
    \subfloat[]{
    \includegraphics[height=0.95in]{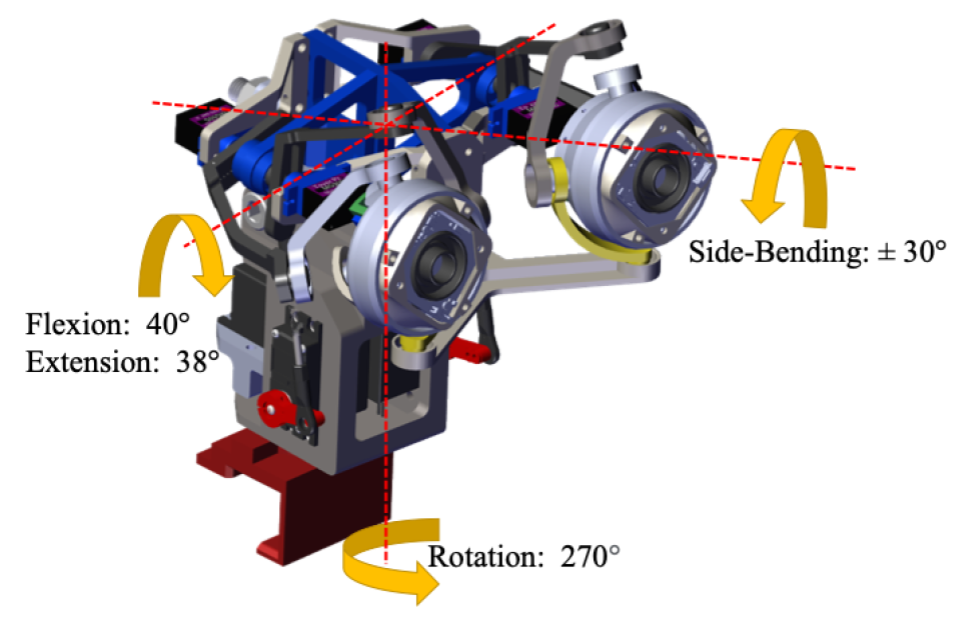}%
    \label{fig_head_dof_full}
    }
    \subfloat[]{
    \includegraphics[height=0.95in]{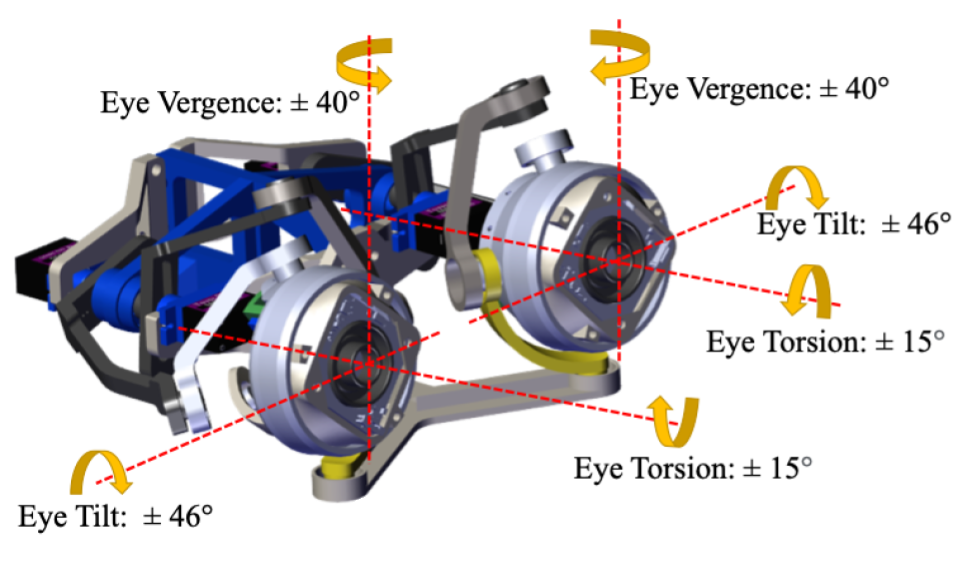}%
    \label{fig_eye_dof_full}
    }
    \caption{Range of motion of DIJIT's head (a) and cameras (b). The range of motion of the human head is roughly $\pm60^\circ$ in yaw (rotation in (a)), $30^\circ$ in flexion, $60^\circ$ in extension, and $\pm20^\circ$ in side-bending~\cite{Head_ROM}. The range of motion of the human eye is roughly $\pm45^\circ$ in vergence, $28^\circ$ in elevation, $47^\circ$ in depression~\cite{Lee2019-rz}, and $\pm10^\circ$ in torsion~\cite{eye-torsion-2}.}
    \label{fig_dof_full}
\end{figure}

\begin{figure}[!t]
    \centering
    \subfloat[]{
    \includegraphics[height=1in]{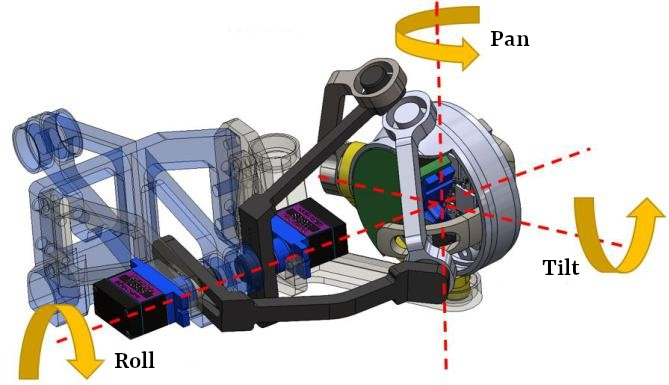}%
    \label{fig_eye_dof_back}
    }
    \subfloat[]{
    \includegraphics[height=1in]{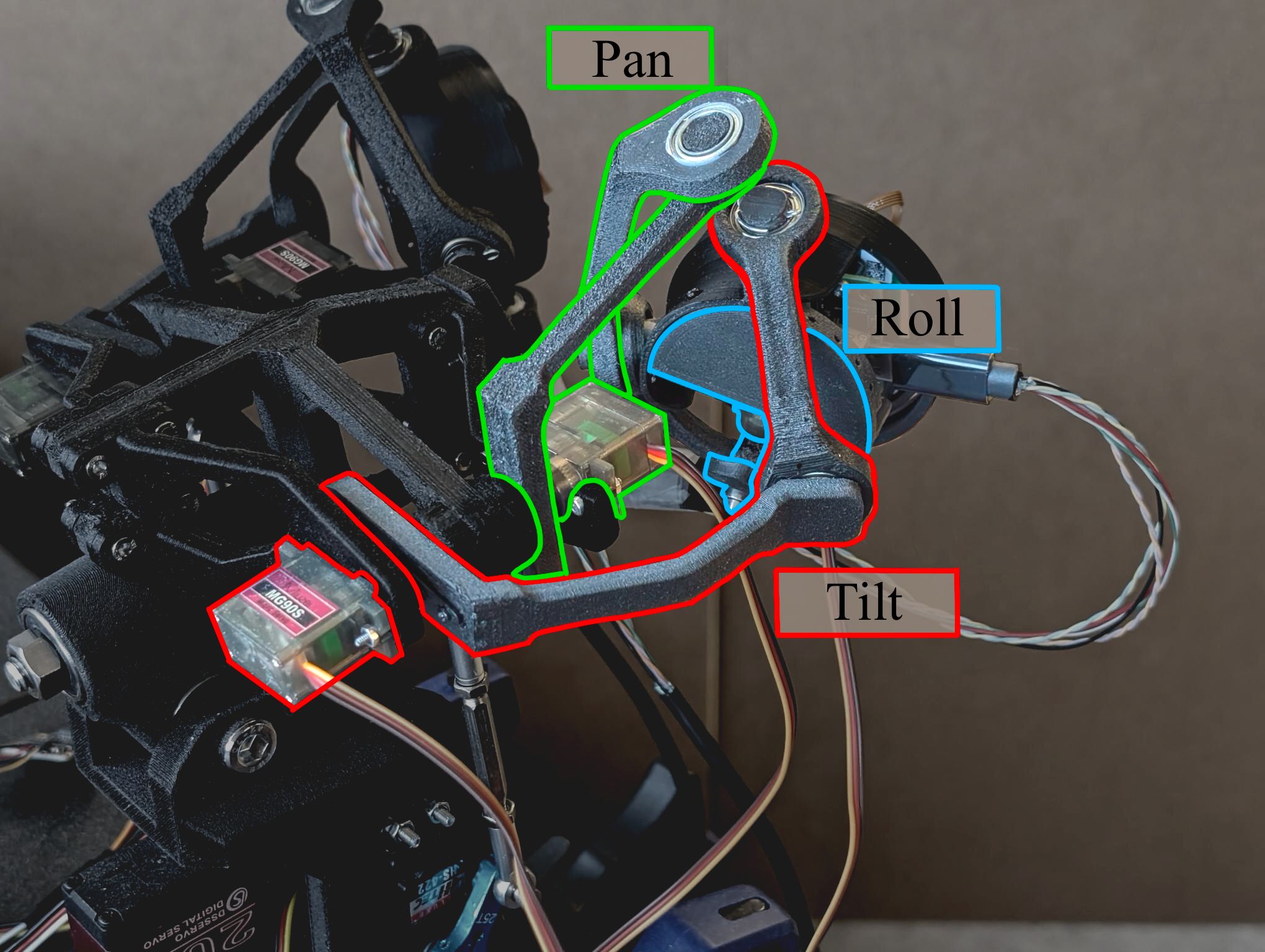}%
    \label{fig_linkages}
    }
    \caption{(a) The DOF of the right camera.
    (b) Red boundaries highlight the motor and linkages responsible for the primary tilting motion, green boundaries highlight those for the primary pan motion, and blue boundaries highlight those responsible for the roll motion.}
    \label{fig_back_linkages}
\end{figure}

\noindent 
DIJIT has nine mechanical DOF: three for the neck, which are pan rotation, side-bending, and flexion/extension; and three for each camera, which are pan, tilt, and roll [See Figs.~\ref{fig_dof_full} and \ref{fig_eye_dof_back}].
The range of motion of each rotation is presented in Fig.~\ref{fig_dof_full}. 
The ranges of motion of the human head and eyes are given in Fig.~\ref{fig_dof_full}.
There are also four optical DOF associated with each camera and lens: focusing distance, camera exposure time, camera white balance, and sensor gain.
Focusing distance is similar to eye accommodation in humans, camera exposure time is similar to visual integration time, camera white balance is similar to color constancy in the visual cortex, and sensor gain is similar to photoreceptor sensitivity.

Three motors control the neck.
A large motor located at the base of the neck controls the pan rotation of the entire head.
Two smaller motors, mounted on either side of the neck, work together to push and pull metal rods on the sides of the neck.
When they pull or push in the same direction, the neck performs flexion or extension, respectively; When they push or pull in opposite directions, a side-bending motion is performed.
Just as the roll, pitch, and yaw rotational motions of the individual camera mirror the rotational motions of the human eye, the neck motions of DIJIT were designed to match the motions of the human neck.

Within the head, the orientation of each camera is controlled by three motors. 
Two motors located behind the camera provide pan and tilt motions. 
Each of them is connected to the camera frame through two connected linkages. 
One linkage is connected to the motor on one side and to the other linkage on the other side. 
The other linkage is connected to the outer surface of the circular camera frame.

To feature all DOF exhibited in the human eye, DIJIT's cameras utilize a third motor responsible for roll that rotates the camera around the line of sight, or principal axis. 
This motor is located behind the camera and is attached to the camera frame. 
There is a ring slightly smaller than the camera frame, sliding inside it, rolling on ball bearings located between them. 
The roll motor is connected to this inner frame through two linkages. 
Fig.~\ref{fig_linkages} highlights these motors and linkages to distinguish between pairs of linkages and their role in the motion of the camera.

DIJIT utilizes a rigid body linkage system, allowing the placement of the motors behind the cameras. 
Although many previous designs align motors with the rotational axes for simpler kinematic calculations, our design allows for a more compact design, with the camera frames being closer, having a baseline length comparable to that of humans.
DIJIT weighs approximately 1.4kg.
The upper part of the head, excluding the neck, is 22cm$\times$18cm$\times$12cm (W$\times$D$\times$H), and the whole head, including the neck, is 22cm$\times$22cm$\times$26cm.
The neck provides a standard mount point for the head to be mounted on a mobile robot platform or to a rigid mount.
For the experiments reported here, a rigid mount was used.

\subsection{Hardware}

\noindent
Important considerations for our design included developing a compact, light, and agile binocular robotic head.
The body, frames, and joint parts are 3D printed using PA6-CF, which is a lightweight, high-performance thermoplastic material.
This is to be contrasted with many other robotic heads, which are constructed using heavy metal.
Each camera is controlled by three MG90S metal gear servos, which are small and light.
The motor located at the base of the neck is a 20kg DC4.8-6.8V servo.
The two motors on the sides of the neck are Hitec HS-422 servos. 
The nine motors are controlled by a Teensy 4.0 ARM microcontroller operating at an I2C data rate of 400~kHz.
The motors are connected to an I2C servo driver interface, which is monitored by the Teensy 4.0. 

The high-performance iDS U3-3881LE-AF camera with a global shutter is used for image capture.
It is powered by and communicates through a single USB 3.0 port at 5 Gb/s for a capture speed of 59 fps at the highest resolution of 3088$\times$2076 pixels. 
The software for the cameras provides control over focusing distance, exposure, white balance, and gain.
Each camera is equipped with a Corning Varioptic C-S-25H0-096 liquid 9.6mm lens with a focus control board that includes an auto-focus feature.
Equipped with these lenses, each camera in DIJIT has a field of view of approximately 40$^\circ\times$30$^\circ$ (H$\times$V).

DIJIT is also equipped with two IMUs that are rigidly connected to the camera frames. Using the BNO085 IMU, the rotational states of the cameras are obtained at 60~Hz.
The states of the IMUs are monitored by the same Teensy that provides motor control.

\subsection{ROS Interface}

\noindent 
The master computer, which runs ROS, communicates with a Teensy 4.0 microcontroller, which controls the motors of the head through serial communications with integers and floats, which represent motor and IMU values, respectively.
Communication between the master computer and the Teensy 4.0 operates at 60~Hz.
Safety check nodes are integrated to check for servo value limit violations before communicating each new motor value sequentially to the Teensy 4.0.
The Teensy 4.0 microcontroller also prompts responses from IMU units and sends signals to the analog servo motors at 60~Hz.
Each camera's current state with IMU values and motor position values, as well as their images, is published at 60~Hz.
Through ROS services, fixations can be changed, motors can be moved and reset, and images can be captured.
The camera parameters, such as focus, gain, white balance, and exposure time, are controllable as well, although we currently set these parameters before startup.

\section{DIJIT Saccade Control} \label{sec:saccade}

\noindent 
Humans are capable of saccade shifts of speeds up to \SI{700}{\degree / \second}~\cite{zuberPeakVelocity}.
A saccade shifts the fixation of a camera from one point to another.
A saccade consists of pure eye motion with no head motion.
When a camera is fixated on a point, that point is projected on the principal point of the camera.
Similar to other saccade methods, we assume the principal point is located at the image center.
In a saccade, the target point is determined and the amount of movement is calculated, and then the camera moves rapidly toward the target.
Methods for determining the next fixation point are outside the scope of this paper. 
We assume that a target point is given in image coordinates relative to one of a set of calibration images, as described later.

\begin{figure}[!t]
    \centering
    \subfloat[]{\includegraphics[height=0.9in]{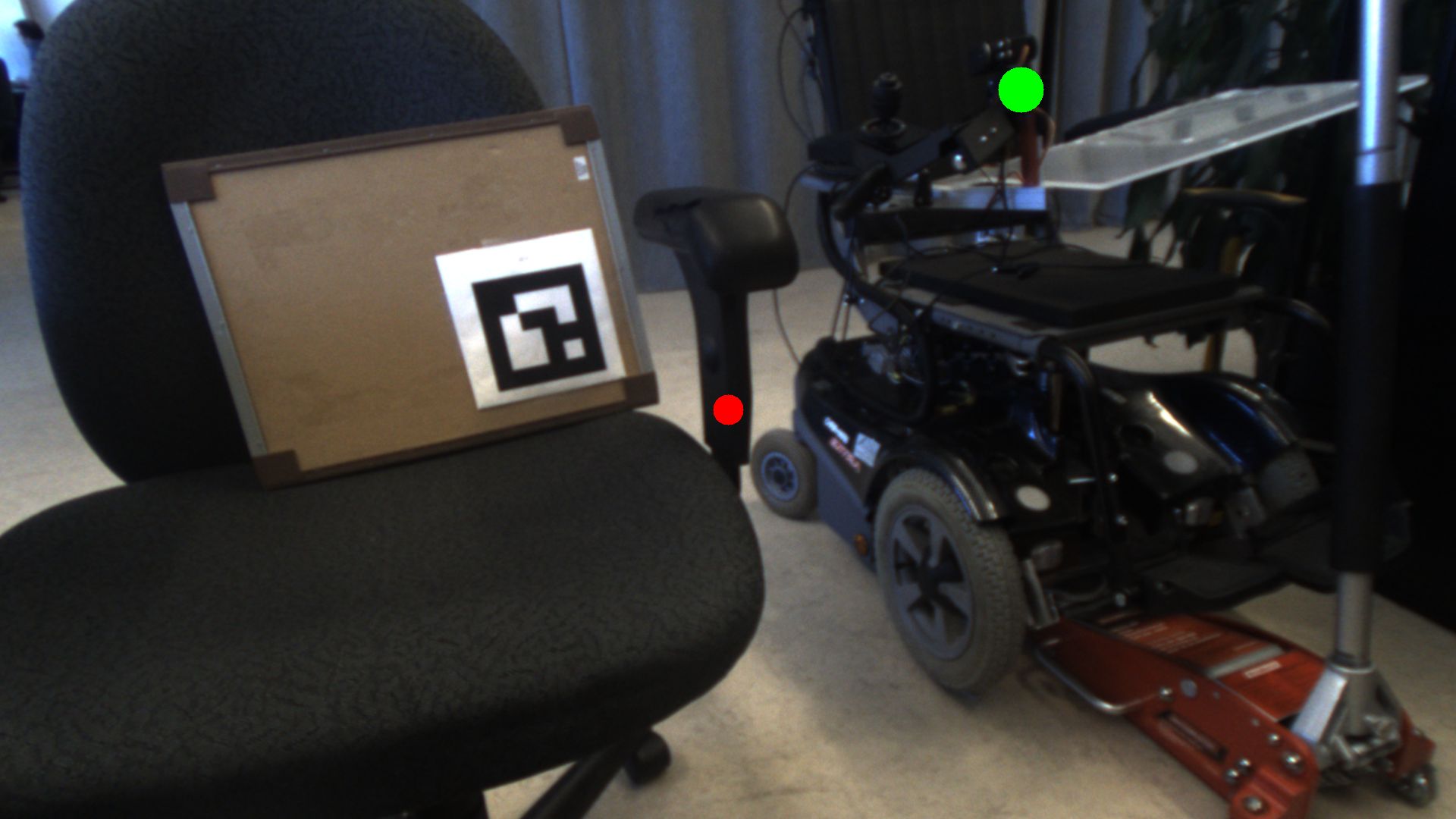}%
    \label{fig_before_sac}}
    \hspace{0.5mm}
    \subfloat[]{\includegraphics[height=0.9in]{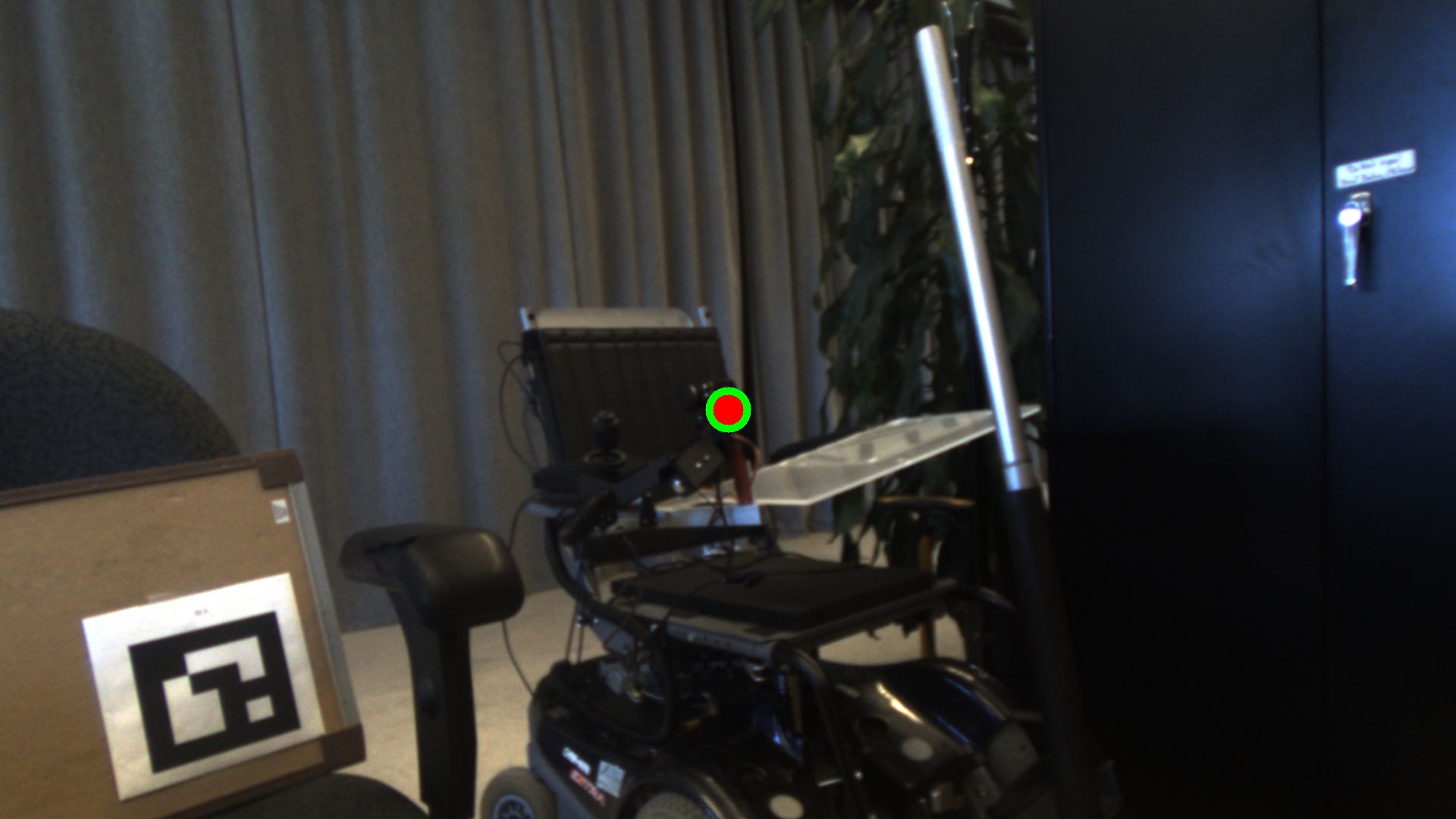}%
    \label{fig_after_sac}}
    \caption{Sample views from DIJIT: The red point indicates the image center in both images.
    (a) Before saccade execution, the green point indicates the target point for fixation.
    (b) After a precise saccade execution, the image is centered at the intended fixation point.}
    \label{fig_sac}
\end{figure}

Fig.~\ref{fig_sac} shows the start and end points of a saccade.
Fig.~\ref{fig_before_sac} shows the state before saccade execution.
The red point is the image center, and the green point is the target point where we wish to fixate.
Fig.~\ref{fig_after_sac}, centered at the green point, shows the state after a precise saccade execution.

What makes this a difficult task to solve is that, in practice, it is difficult to construct a device that rotates the camera about the camera's optical center.
Difficult to estimate offsets between the motor axes and the optical ones complicates the accurate estimation of the appropriate motor rotations. 
Here, we propose a data-driven method to avoid explicitly modelling and calibrating the relationship between the motor rotational axes and the change in saccade direction. 

Our approach relies on a set of calibration views of a calibration target that enables estimation of a homography between the views of any two images in the calibration set.
This allows us to map the center of each calibration image to the calibration image corresponding to the pixel location of the planned saccade.
Since not all motor states are recorded in the calibration data collection, some interpolation process is required to approximate unseen views.
The set of calibration mappings is used along with the known pan and tilt values of the calibration images to estimate the pan and tilt values required for the saccade.
This process is applied separately to the left and right cameras using the pan and tilt motions needed for a simple saccade. 
It can easily generalize to cyclotorsion, which may be additionally needed for off-horopter binocular fusion (ongoing work).

\subsection{Calibration Data Collection} \label{sec_data_collection}

\noindent
Rather than relying on modelling the kinematic structure of the camera pan, tilt, and roll motors, we address the problem by assuming an idealized model of the kinematics with rotation about the camera center and utilizing a standard visual target (ChArUco~\cite{garrido-juradoAutomaticGenerationDetection2014}) to link camera motions to pixel offsets. 
We start by collecting a set of calibration images to develop the relationship between potential fixation points and motor settings for the pan and tilt motors.
A ChArUco board is fixed in front of DIJIT.
The pan and tilt motors are moved in discrete steps (e.g., five degrees) over the range of the pan and tilt values, and images are captured of the target. 
These pan and tilt values are chosen to provide a dense sampling of the pan and tilt space.
Note that here we assume that the same calibration target is visible in all views, although the approach can be easily generalized using multiple ChArUco targets in different positions.

When the pan and tilt motors are in positions $p_i$ and $t_i$, respectively, the motor values $\left(p_i, t_i \right)$ and the image $I_{i}$ taken in this state are recorded.
The IMU values $\left(s_{i}^x, s_{i}^y, s_{i}^z\right)$, which are three Euler angles, are recorded as well, although these values are not used in the work described here.
This results in a set of calibration data, $\mathcal{C} = \{(I_i, p_i, t_i, s^x_i, s^y_i, s^z_i)\}$.

When a camera undergoes pure rotation, a homography relates two images taken by it.
This homography only depends on the camera rotation; it is independent of the structure and other properties of the environment.
Therefore, a homography estimated for images taken of the calibration object holds even for images taken in the same motor states in other environments with different properties.
Factors like partial occlusion and light variations do not affect our calibration; instead, they may affect detecting a new interesting point to fixate on.
This is out of the scope of this paper. 
We assume a target point is given.

Every time we disassemble and reassemble a camera or its motors, that camera must be recalibrated.
Otherwise, it can operate for hundreds of saccades over multiple days without requiring recalibration.

\subsection{Fixation Point to Motor State Correspondence} \label{first_saccade_method}

\noindent
Assume that for all possible motor states, the corresponding data is recorded during the calibration data collection stage.
Let $\mathbf{t} = [t_x, t_y]^T$ be the image point where we wish to fixate, defined in image $I_s$, an image in the calibration set $\mathcal{C}$, which is the starting direction of the saccade.
Point $\mathbf{t}$ is initially obtained in the current view of the camera at the test time and then transferred to image $I_s$ in the same state of the motors in the calibration set $\mathcal{C}$.
The methods to obtain point $\mathbf{t}$ are beyond the scope of this paper, and we assume this point is given, see Section~\ref{experiments}. 
For sample $k$ in the calibration set, we establish the homography between $I_k$ and $I_s$ and use this to estimate $\mathbf{c}_k$, the location of the center of image $I_k$ as viewed in $I_s$.
We define an image-space error metric between $\mathbf{c}_k$ and $\mathbf{t}$ to define the closest $\mathbf{c}_k$ in $I_s$.
Here, we use the Cartesian distance in image space as this metric.
Since the pan and tilt motors are moved in discrete steps (e.g., five degrees) in the data collection stage, the closest $\mathbf{c}_k$ may correspond to an unseen view.
Thus, we utilize a bilinear interpolation process to estimate $\mathbf{c}_k$ for unseen views, see Section~\ref{sec_interpolation}.

\subsection{Finding Correspondences of Image Centers} \label{sec_cam_motor_calib}

\noindent 
We aim to determine point $\mathbf{c}_k$ in image $I_{s}$, which corresponds to the center of image $I_k$.
We accomplish this by estimating a homography from image $I_k$ to image $I_s$.
A homography relates two images in two cases: when the captured scene is a planar surface in 3D space or when the camera undergoes pure rotation~\cite{hartley_multiple_2004}.

The ChArUco board is visible in both images and forms a planar region in them, enabling us to estimate a homography that maps between them.
Let $\{ \mathbf{a}_1, \mathbf{a}_2, ..., \mathbf{a}_r \}$ and $\{ \mathbf{b}_1, \mathbf{b}_2, ..., \mathbf{b}_r \}$ be corresponding corners of the ChArUco board in the images $I_{s}$ and $I_{k}$, respectively.
We use these correspondences to estimate the homography matrix $\mathbf{H}_k$ from image $I_k$ to image $I_s$.
Thus, we have:
\begin{equation}
    \overline{\mathbf{a}}_j = \mathbf{H}_k \ \overline{\mathbf{b}}_j \quad \textrm{for} \ j=1, 2, ..., r
\end{equation}
where $\overline{\mathbf{a}}_j$ and $\overline{\mathbf{b}}_j$ are the homogeneous coordinates of $\mathbf{a}_j$ and $\mathbf{b}_j$, respectively.

We assume that the camera rotates around its center; therefore, although this homography is obtained for a planar surface covering only part of the images, it relates all pixels between the two images.
Thus, we apply this homography to the image center $\mathbf{p}$ in $I_k$ to determine its corresponding point $\mathbf{c}_k$ in $I_s$:
\begin{equation} \label{homography}
    \overline{\mathbf{c}}_{k} = \mathbf{H}_k \ \overline{\mathbf{p}}
\end{equation}
\noindent where $\overline{\mathbf{c}}_{k}$ and $\overline{\mathbf{p}}$ are homogeneous coordinates of $\mathbf{c_{k}}$ and $\mathbf{p}$, respectively.

\subsection{Interpolation of Unseen Motor States} \label{sec_interpolation}

\noindent 
Since the pan and tilt motors are moved in discrete steps (e.g., five degrees), using the same notation as in the previous sections, image $I_s$ or $I_k$ or both may not be among the sampled data.
In such a case, we use bilinear interpolation to find $c_k$.
When $I_s$ is not available, and $I_k$ is available, we find $c_k$ in the four closest collected images to $I_s$, and interpolate them bilinearly to find $c_k$ in $I_s$.
When $I_s$ is available, and $I_k$ is not, we find the four closest collected images to $I_k$, find the corresponding point to the image center of each of them in $I_s$, and bilinearly interpolate them to find $c_k$ in $I_s$.
When neither $I_s$ nor $I_k$ are vailable in the collected data, we combine the two bilinear interpolations above.

\begin{table}[!t]
    \renewcommand{\arraystretch}{1.3}
    \caption{Performance Evaluation
    \label{tab:errors}}
    \centering
    \begin{tabular}{l|llll}
    \hline
    
            & \multirow{2}{*}{Saccade}  & Saccade   & Motor         & Assumes       \\
    Method  & \multirow{2}{*}{Error}    & Speed     & Resolution    & Precise       \\
            &                           & (deg/sec) & (degree)      & Robot$^{(1)}$ \\  
    
    \hline
    
    \multirow{2}{*}{Human}  & \multirow{2}{*}{$^{(2)}$} &   Up to 700               & \multirow{2}{*}{N/A$^{(3)}$}  & \multirow{2}{*}{N/A}  \\
                            &                           & ~\cite{zuberPeakVelocity} &                               &                       \\
    
    \hline
    
    DIJIT                                                           & 1.17$^\circ$ - 1.14$^\circ$   & Up to 600             & 1                     & No            \\
    
    \hline
    
    \textit{Manfredi$^{(4)}$~\cite{manfredi_implementation_2006}}   & \textit{1.57}$^\circ$         & \textit{Up to 120}    & \textit{N/R}$^{(5)}$  & \textit{Yes}  \\
    
    \hline
    
    \textit{Van Opstal~\cite{vanOpstal}}                            & \textit{1.47}$^\circ$         & \textit{344}          & \textit{N/R}          & \textit{Yes}  \\
    
    \hline
    
    iCub~\cite{muhammad_neural_2015}                                & $<1^\circ$                    & Up to 50              & 0.005                 & Yes           \\
    
    \hline
    
    COG~\cite{goos_cog_1999}                                        & $<1$ pixel                    & Up to 360             & 0.125                 & No            \\
    
    \hline
    
    Tombatossals~\cite{antonelli_learning_2015}                     & $>5$ pixels                   & N/R                   & N/R                   & Yes           \\
    
    \hline
    
    Schenck~\cite{schenck_robot_2013}                               & 1 pixel                       & N/R                   & N/R                   & Yes           \\
    
    \hline
    
    OREO~\cite{huber_oreo_2018}                                     & N/R                           & Up to 535             & 0.036                 & Yes           \\
    
    \hline
    \end{tabular}
    \begin{flushleft}
    \footnotesize
    $^{(1)}$~``Assumes Precise Robot'' means whether the method assumes the alignment of the camera, eye frame, and motors.
    $^{(2)}$~Reported human saccade errors vary from 0.6$^\circ$ to 2.89$^\circ$ depending on different factors, such as the target eccentricity, saccade direction, and task and experiment conditions~\cite{endPointError, ZimmermannSaccade, hanningVisualAttentionNot2019, zivotofskySaccadesRememberedTargets1996}.
    $^{(3)}$~N/A stands for Not Applicable.
    $^{(4)}$~Methods shown in \textit{italic} have reported errors only in simulation.
    $^{(5)}$~N/R stands for Not Reported. 
    \end{flushleft}

\end{table}

\section{Experimental Results} \label{experiments}

\noindent
We evaluate the saccade method using both the left and the right cameras of DIJIT.
To conform with the other methods to which we compare, Table~\ref{tab:errors}, we assume that each camera is fixated on its image center.
In the experiments we conducted, an ArUco marker was placed in front of the head in the field of view of both cameras, its center was detected using OpenCV in each camera as the target point, and both cameras were moved to fixate on the same ArUco marker.
In a saccade movement, no information about the target point is updated from sending a saccade execution command until its completion.
A precise saccade should land in the target location detected at the time of sending the execution command, and this is the point of our current tests.
Naturally, in real scenes, targets may move, and to detect such motion, some visual processing is needed, and this takes time (about 100~ms or so in humans).
Thus, any reaction, whose determination also needs additional time, would only be known during the already executing saccade. 
This additional visual processing, decision-making, and control are not in the scope of the current work but are topics of our active research.

Once the binocular saccade mechanism is executed, it takes 12~ms to compute the new motor values for all four motors, two motors for each camera, using an AMD Ryzen 7 7700X 8-Core processor in the master computer.
Then, the computed values for the two motors of the left camera, followed by the two motors of the right camera, are sequentially sent to the Teensy 4.0 microcontroller at 60~Hz, and the Teensy board sends these values to a PWM servo driver connected to the servo motors at 60~Hz. 
The right camera starts moving 33~ms after the left camera.
After executing a saccade, the center of the ArUco marker is detected again for performance evaluation.

To plan the next saccade, the location of the ArUco marker was changed, and it was located in a new position in the field of view of both cameras.
We conducted 175 trials in total, performing saccades in both cameras together in each trial.
To include different scenarios, in 91 experiments, the cameras moved back to their home positions, where they were looking straight ahead, and started the next saccade from there.
In the other 84 experiments, the start direction of the next saccade was the end direction of the current saccade.
The placement of the ArUco marker was sampled such that the distribution of the target eccentricities would be similar to that of humans.
Specifically, 58\% of the saccades were within 6$^\circ$ of the image center, 27\% between 6$^\circ$ and 12$^\circ$, and 15\% larger than 12$^\circ$.
The largest saccades were 19.08$^\circ$ and 18.41$^\circ$ in the left and right cameras, respectively.
This distribution approximates the distribution of human saccade amplitudes, as described in~\cite{tatler_long_2006}. 
The amplitude of a saccade is the angular distance between the target point and the image center.

Saccade error is defined as the angle between the back-projected ray of the landing point and that of the image center.
A landing point is the location of the target point after executing the saccade, which should be at the image center.
The average saccadic error was $1.17^\circ\pm0.86^\circ$ and $1.14^\circ\pm0.67^\circ$ (mean$\pm$std) for the left and right cameras, respectively, which are acceptable values compared to human saccade error.
Different values are reported for human saccade errors in the literature depending on various factors, such as the target eccentricity, saccade direction, and task and experiment conditions.
In~\cite{endPointError}, the error for saccades of amplitude 10$^\circ$ is reported as 2.89$^\circ$ when a distractor is present and 1.17$^\circ$ in the absence of a distractor.
In~\cite{ZimmermannSaccade}, the reported error for 8$^\circ$ rightward horizontal saccades is 1.2$^\circ$ for long latency and 0.6$^\circ$ for short latency.
In~\cite{hanningVisualAttentionNot2019}, for saccades of amplitude 10$^\circ$ leftward horizontally, the error is reported as 1.25$^\circ$, for 6$^\circ$ leftward it is 0.79$^\circ$, and for 6$^\circ$ rightward it is 1.11$^\circ$.
Note that human saccades are most accurate along the horizontal direction.
In~\cite{zivotofskySaccadesRememberedTargets1996}, the median saccadic error for flashed targets is 0.93$^\circ$ horizontally and 1.1$^\circ$ vertically, and for visible targets it is 0.59$^\circ$ horizontally and 0.60$^\circ$ vertically.
The median error of our method is 0.73$^\circ$ and 0.80$^\circ$ horizontally and 0.29$^\circ$ and 0.37$^\circ$ vertically for the left and right cameras, respectively.

In Table~\ref{tab:errors}, we compare our results with previously reported saccade results from the literature.
Some methods~\cite{vanOpstal, muhammad_neural_2015} are only evaluated in simulation.
Simulations may involve many simplifications and may not account for mechanical and physical limitations and for deviations when manufacturing and assembling the robot; therefore, the results may vary significantly when applied to real robots.
In these simulations, the error reported in~\cite{vanOpstal} is 1.47$^\circ$; and while the mean error reported in~\cite{muhammad_neural_2015} is less than 1$^\circ$, the error was calculated after a corrective saccade.
DIJIT's error after a corrective saccade is 0.93$^\circ$ and 0.95$^\circ$ for the left and right cameras, respectively.

For~\cite{goos_cog_1999},~\cite{antonelli_learning_2015}, and~\cite{schenck_robot_2013}, the saccade error is reported in pixels, not in angular terms. 
Pixel error is not an ideal measure for estimating error because its significance varies depending on the pixel size of the camera sensor and the focal length of the lens. 
We converted these pixel errors into angular errors using the camera and lens information provided by the authors. 
In~\cite{goos_cog_1999}, where less than one pixel of error in fixation is reported, the images are 120$\times$120 pixels, and the field of view is 115.8$^\circ$$\times$88.6$^\circ$.
Based on these values, one pixel distance from the image center is between 0.93$^\circ$ and 1.52$^\circ$, which is comparable to our results. 
The method in \cite{schenck_robot_2013} operates on images resembling retinal images and reports an error equivalent to 1.8$\%$ of the size of a plain image or one pixel in an image of size 100. 
We were not able to convert this value into an equivalent angular error based on the information provided in their paper. 
In \cite{antonelli_learning_2015}, the authors report that their best performance among different parameters was 0.47 and 5 pixels of error for binocular and monocular cases, respectively, in simulation; and 3.70 pixels of error for the binocular case in the real robot.
They do not report the monocular result for the real robot; however, they report that the pixel size of their camera sensor in the real robot is 4.65$\mu$m, and the focal length of the lens is 5mm.
3.70 pixels of binocular error in the real robot is equivalent to 0.2$^\circ$.
Considering a similar relative difference between binocular and monocular performances in simulation and real robot, we expect the monocular result in the real robot not to be better than our result.

The maximum saccade speed of DIJIT surpasses all other reported values in Table~\ref{tab:errors}. 
Although the motor encoders used in DIJIT do not have as high a resolution as encoders used in other biocular heads, we obtain better or comparable accuracies in executing saccades. 
This indicates the strong performance of DIJIT and the accuracy of the saccade method. 

On average, the error for saccades with smaller amplitudes was less than that for larger amplitudes measured in degrees; however, in the right camera, the average error of very large saccades was smaller than the average error of medium amplitude saccades.
Specifically, the average error for saccades with an amplitude less than 6$^\circ$ was 1.00$^\circ$ and 0.92$^\circ$, for saccades with an amplitude between 6$^\circ$ and 12$^\circ$, it was 1.33$^\circ$ and 1.50$^\circ$, and for large saccades with an amplitude larger than 12$^\circ$ it was 1.56$^\circ$ and 1.28$^\circ$, for the left and right cameras, respectively. 

The distribution of the target's eccentricity in our experiments is very similar to that in humans.
When humans are asked to freely look at images displayed on a monitor~\cite{tatler_long_2006}, 52$\%$ of the saccades are between 0$^\circ$ and 6$^\circ$, 33$\%$ are between 6$^\circ$ and 12$^\circ$, 11$\%$ are between 12$^\circ$ and 18$^\circ$, and 4$\%$ are larger than 18$^\circ$.
DIJIT's performance has a mean performance comparable to that of humans.
After conducting a corrective saccade, the average error for DIJIT is decreased to 0.93$^\circ$ and 0.95$^\circ$ for the left and right cameras, respectively (0.82$^\circ$ and 0.76$^\circ$ for 0$^\circ$-6$^\circ$, 0.98$^\circ$ and 1.35$^\circ$ for 6$^\circ$-12$^\circ$, and 1.32$^\circ$ and 1.16$^\circ$ for larger than 12$^\circ$).
We perform a corrective saccade only when the error of the primary saccade is larger than 1$^\circ$.

We also considered fixation errors in the horizontal and vertical directions. 
Consider a 2D coordinate system on the image where its center is the image center, and the axes are aligned with the horizontal and vertical borders of the image.
We projected the landing points onto these two axes and calculated the angle between the back-projected rays of these projections and the image center. 
The average horizontal and vertical errors in the 175 experiments were $1.03^\circ\pm0.88^\circ$ and $0.40^\circ\pm0.36^\circ$, respectively, for the left camera, and $0.92^\circ\pm0.69^\circ$ and $0.49^\circ\pm0.40^\circ$ for the right camera.

\begin{figure}[!t]
    \centering
    \subfloat[]{\includegraphics[height=1.25in]{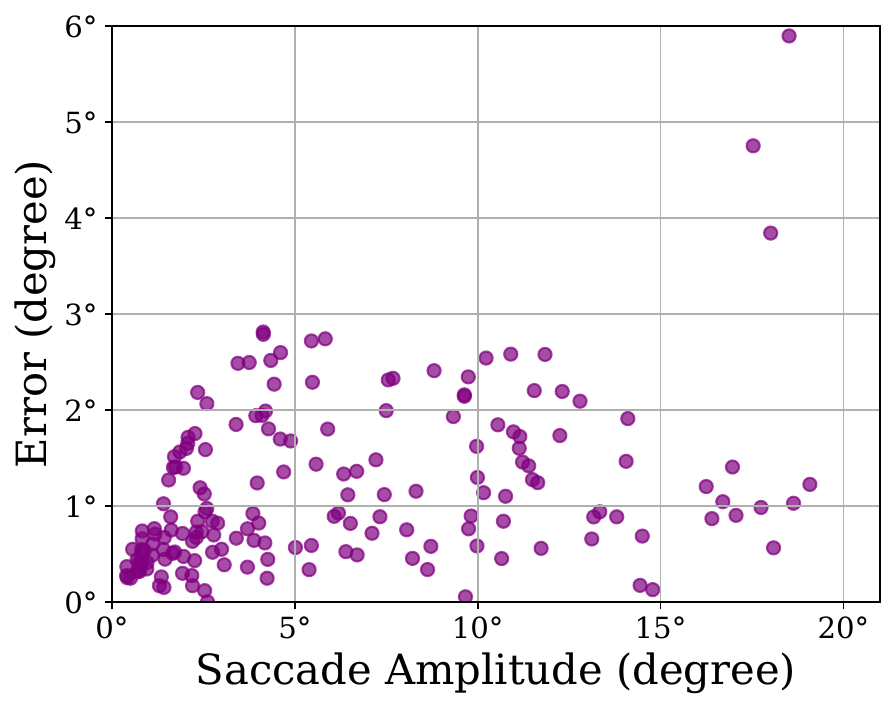}%
    \label{fig_large_errs}}
    \hspace{1mm}
    \subfloat[]{\includegraphics[height=1.25in]{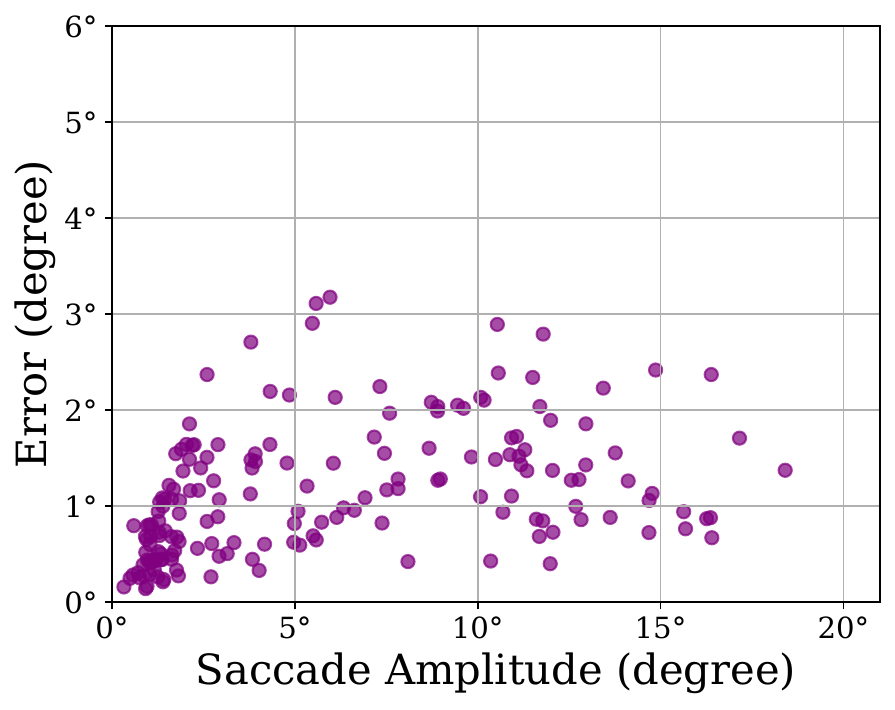}%
    \label{fig_small_errs}}
    \caption{
    Saccade Amplitude versus Error.
    Each point represents a trial.
    The horizontal axis is the saccade amplitude, and the vertical axis is the error.
    (a) Left camera.
    (b) Right camera.
    }
    \label{fig_errs}
\end{figure}

Over 175 experiments, for $55\%$ and $49\%$ of saccades, for the left and right camera, respectively, the error after completing the primary saccade is less than 1$^\circ$, and for $85\%$ and $87\%$ of saccades, the error is less than 2$^\circ$.
Fig.~\ref{fig_errs} illustrates amplitude versus error for all 175 experiments in both cameras.
Each point corresponds to a trial, the horizontal axis is the amplitude of the saccade in that trial, and the vertical axis is the error.

To test the repeatability, we located the ArUco marker in ten different positions, and for each of them, we repeated a binocular saccade ten times, starting from the home position of the cameras.
The average saccade amplitude in all 100 trials was 11.67$^\circ$ and 11.25$^\circ$, and the average saccade error was 1.13$^\circ$ and 1.43$^\circ$, for the left and right cameras, respectively.
For each ArUco marker position, we verified whether the target was located in the same coordinates each time the cameras moved back to their home positions.
For each ArUco position, we computed the standard deviation (SD) of the target eccentricity along the horizontal and vertical axes across ten trials, and then averaged these values over all ten ArUco positions.
The SD for the left camera was 0.3$^\circ$ and 0.07$^\circ$ for horizontal and vertical axes, respectively, and those for the right camera were 0.16$^\circ$ and 0.13$^\circ$.
We performed a similar study to verify whether the coordinates of the landing points in ten trials for each ArUco position were similar enough.
We computed the SD of all errors for each ArUco position along the horizontal and vertical axes, and averaged these values.
The SD for the left camera was 0.36$^\circ$ and 0.13$^\circ$ along the horizontal and vertical axes, respectively.
These values were 0.22$^\circ$ and 0.21$^\circ$ for the right camera.

\section{Discussion and Future Work} \label{sec:review}

\noindent 
This paper introduces the DIJIT binocular robotic head. 
To the best of our knowledge, this is the first open-source robotic head with full nine mechanical DOF and four optical DOF with a baseline close to human specification. 
It is well-suited for active vision tasks and for evaluating theories in cognitive science and neuroscience.
It is straightforward and affordable to replicate DIJIT: the design and software are open source, and the hardware and 3D printing parts are inexpensive.

We developed a binocular saccade algorithm to drive the cameras in the head. 
We demonstrate performance similar to that found in humans and further provide a comparison between saccade performance with DIJIT relative to published performance of other robot heads in the literature.

Numerous existing saccade methods typically assume that the robot is highly precise: the camera's coordinate system, the camera frame's rotational axes, and the rotating axes of the motors are all perfectly aligned. 
The method described here does not depend on these assumptions and, as a result, inherently avoids the need for precise axis alignment.
Instead, a data-driven approach is used to map from pan and tilt motor states to image coordinates, coupled with an interpolation process.
Performance results indicate the accuracy of the approach.

Ongoing work on DIJIT includes work on other eye movements, the development of active binocular 3D reconstruction algorithms, and the deployment of the device within an autonomous system to support wheelchair-bound individuals.

\section*{Acknowledgments}

This material is based upon work supported by the Air Force Office of Scientific Research under award number FA9550-22-1-0538 (Computational Cognition and Machine Intelligence portfolio); the Canada Research Chairs Program Grant Number 950-231659; and Natural Sciences and Engineering Research Council of Canada Grant Number RGPIN-2022-04606.
All grants are awarded to John K. Tsotsos.
Funders did not play any role in study design, data collection, analysis, decision to publish, or preparation of the manuscript.

\bibliographystyle{IEEEtran}
\bibliography{DIJIT.bib}

\end{document}